\documentclass{article}

\usepackage{arxiv}

\usepackage{graphicx}
\usepackage{amsmath,amssymb}
\usepackage{url}
\usepackage{multirow}
\usepackage{subcaption}
\usepackage{color}
\usepackage{verbatim}
\usepackage{bm}
\usepackage{bbm}
\usepackage{amsthm}

\newtheorem{theorem}{Theorem}

\title{Towards a General Framework to Embed Advanced Machine Learning in Process Control Systems}

\author{
  Stefan Schrunner\\
  Department of Data Science\\
  Norwegian Univ. of Life Sciences\\
  Ås, Norway \\
  \texttt{stefan.schrunner@nmbu.no} \\
  \And
  Michael Scheiber\\
  KAI GmbH\\
  Villach, Austria\\
  \texttt{michael.scheiber@k-ai.at} \\
  \And
  Anna Jenul\\
  Department of Data Science\\
  Norwegian Univ. of Life Sciences\\
  Ås, Norway \\
  \texttt{anna.jenul@nmbu.no} \\
  \And
  Anja Zernig\\
  KAI GmbH\\
  Villach, Austria\\
  \texttt{anja.zernig@k-ai.at} \\
  \And
  Andre Kaestner\\
  Infineon Technologies Austria AG\\
  Villach, Austria\\
  \texttt{andre.kaestner@infineon.com} \\
  \And
  Roman~Kern\\
  Know-Center GmbH\\
  Graz University of Technology\\
  Graz, Austria\\
  \texttt{rkern@know-center.at}
}

\begin{document}
\maketitle

\begin{abstract}
%Abstract
Since high data volume and complex data formats delivered in modern high-end production environments go beyond the scope of classical process control systems, more advanced tools involving machine learning are required to reliably recognize failure patterns. However, currently, such systems lack a general setup and are only available as application-specific solutions. We propose a process control framework entitled Health Factor for Process Control (HFPC) to bridge the gap between conventional statistical tools and novel machine learning (ML) algorithms. HFPC comprises two main concepts: (a) pattern type to account for qualitative characteristics (error patterns) and (b) intensity to quantify the level of a deviation. While the system retains large model generality, allowing a broad scope of potential application areas, we demonstrate its favorable mathematical properties in a theoretical analysis. In a case study from the semiconductor industry, we underline that (a) our framework is of practical relevance and goes beyond conventional process control, and (b) achieves high-quality experimental results. We conclude that our work contributes to the integration of ML in real-world process control and paves the way to automated decision support in manufacturing.
\end{abstract}

% keywords can be removed

\section{Introduction}
Across all industries, advanced manufacturing processes require permanent monitoring to guarantee the high quality of products. Automation of such processes demands even more autonomous and flexible control algorithms. Meanwhile, larger and more complex data are produced, such that errors detection requires pattern recognition in unstructured and high-dimensional datasets rather than tracking single parameters. Since classical means of process control cannot cover this aspect, not only production but also process control and monitoring require novel concepts.
\subsection{Background}
Classical process control frameworks, like control charts~\cite{shewhart1931} and related statistical approaches, are tailored to analyze the distribution of single (or just a few) parameters over time.
However, their ability to cover data formats produced by manufacturing equipment in Industry 4.0 is limited since modern manufacturing environments deliver sensor data with complex temporal and spatial dependencies (e.g., images).
Thus, machine learning (ML) techniques have become a vital part of real-world automation systems in recent years and opened new possibilities in this area, such as characterizing patterns that point to erroneous states in the process. However, integrating ML methods into process control is not trivial due to fundamental differences from the classical statistical frameworks.

Multiple research works suggest deploying ML algorithms for process control --- many of these are targeted to specific applications in the semiconductor industry~\cite{yu2012,liu2020} or oil industry~\cite{GAN202283}.
Such approaches introduce anomaly detection methodology~\cite{7560618}, graph-theoretic models for time series analysis~\cite{7556960}, or Gaussian mixture models~\cite{yu2012} to improve process control.
More recent works cover a system for detecting incipient faults~\cite{ZHANG2020173}, a hybrid machine learning framework using an extended state-space model~\cite{SUN202030}, and a simulation-based model to learn decision rules~\cite{DOMAHIDI2014763}.
Even though many of these works achieve good results and outperform conventional techniques, most of them are restricted to specific scenarios or problems rather than allowing generalization across problem setups, processes, or even industries.
Furthermore, scientific papers from the machine learning area tend to focus on performance-driven metrics evaluated on test data. At the same time, statistical properties of the suggested process indicators remain unexplored --- even though these aspects are of high importance for practitioners to understand the benefits and shortcomings of their process control system.
Thus, a novel, statistically well-founded framework is required to bridge the gap between the classical and ML approaches in process control.

\subsection{Problem formalization and contribution}

\begin{figure}
    \centering
    \includegraphics[width = 0.45\textwidth]{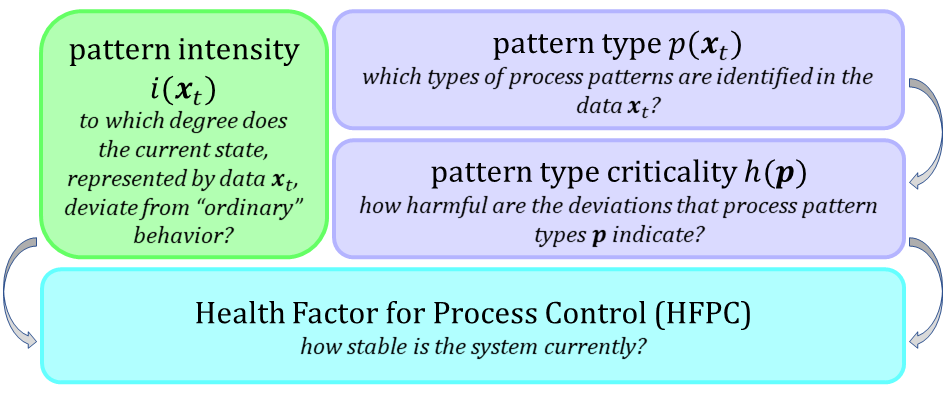}
    \caption{Components of the HFPC. An intensity-based indicator reflecting the degree of a deviation is combined with data-driven ML models to detect pattern types. Expert knowledge is required to assess if a pattern is considered critical for the product quality.}
    \label{fig:concept}
\end{figure}

To provide a general framework for machine learning based process control, we define an indicator for process deviations, which is a generalization of our previous work~\cite{schrunner19}: the Health Factor for Process Control (HFPC), denoted as $\text{HF}(\bm{x}_t)$ for some observation $\bm{x}_t\in \mathcal{D}$ from the process at a time $t$, where $\mathcal{D}$ represents some feature space. Observations may represent sensor data or intermediate tests. As sketched in Fig. \ref{fig:concept}, our concept of the HFPC involves two independent components:
\begin{itemize}
    \item a function $p$, which maps the observation to the binary vector indicating detected pattern types, $p:\mathcal{D}\rightarrow \{0,1\}^{K}$, where $K=\vert \mathcal{P}\vert$ denotes the cardinality of the set of all known process patterns $\mathcal{P}$, along with pattern type criticality levels $h:\{0,1\}^{K}\rightarrow [0,1]$, and
    \item an intensity measure $i:\mathcal{D}\rightarrow [0,1]$.
\end{itemize}
An arbitrary pattern recognition process determines the pattern type with the ability to discriminate between types of process patterns (e.g., distinguishable classes of shapes). Pattern types are linked to a criticality value, a level of potential quality concern associated with the occurrence of the patterns. Independently, the concept of intensity describes the different degrees of deviations (ranging from weak to very strong states). In a weak state, patterns slightly deviate from measurement noise. In contrast, in a strong state, deviations are large and may scratch the specification limits.\footnote{Note that the concepts of pattern type and intensity do not overlap. While pattern type describes abstract properties like shapes or positions of anomalies relating to different types of failures, intensity describes the degree of development of deviations from ordinary states or random noise.}

In order to retain full generality of our framework, implementations of the generic components $i$, $p$, and $h$ may be automated, semi-automated or manual. In a generic way, we treat all of these quantities as random variables, following a probability distribution over the according function domain, e.g. over all combinations of process patterns, given as $P(p(\bm{x}_t)=\bm{p})$ for $\bm{p}\in\{0,1\}^K$, for the pattern type classifier $p$. However, deterministic methods may still be used by specifying one-point distributions. In the following we assume $h$ to be derived from expert knowledge, while $p$ and $i$ represents a data-driven components involving ML methods.

To preserve model generality, we define $\text{HF}(\bm{x}_t)$ as a product of the presented model components:
\begin{equation}
\text{HF}(\bm{x}_t) = i(\bm{x}_t) \cdot h(p(\bm{x}_t)),
\label{eq:hf}
\end{equation}
In practice, model components $i$, $p$ and $h$ are replaced by the expected values of their respective (problem-specific) estimators.
Based on the definition of $\text{HF}$ our main contribution in this work is to (a) investigate the mathematical and statistical properties of the HFPC as a general process monitoring framework to integrate machine learning components and (b) demonstrate the capabilities of the concept in a practical use-case. Specifically, we analyze the compatibility of HFPCs with established process control mechanisms and derive expressions for the expected false positive/false negative rates of a decision taken based on HFPC.
Our experimental part demonstrates the concept's capabilities in a case study from the semiconductor industry. The detection of process patterns goes beyond assessing classical pass/fail test investigations by using methods from image processing and ML in analog wafer test data.

\section{Theory}
\label{sec:theory}

We first develop the theory behind the concept of the HFPC, $\text{HF}(\bm{x}_t)$, given in Eq. \ref{eq:hf}. For this purpose, we state the model assumptions underlying the concept, followed by an analysis of statistical properties, which aid the user in understanding, interpreting, and assessing the indicator in practice.

\subsection{Model assumptions}

% pseudometric
Given the definition of the $\text{HF}(\bm{x}_t)$, the function $\text{HF}:\mathcal{D} \rightarrow \mathbb{R}$ assigns a level of concern, risk or loss to any data point $\bm{x}_t$. 
By its definition, the HFPC is \textbf{non-negative} if both intensity $i$ and criticality $h$ are non-negative, which is a common and intuitive assumption for such quantifiers. An upper bound to the HFPC is given by the product of the upper bounds of $i$ and $h$, in case both exist. For practical applications, we assume that $i$ and $h$ are bounded above by 1 to facilitate interpretation, which can be achieved for arbitrary finite variables by monotonic transformations.

% availability of labeled training data - rephrase for patterns vs criticality
If not provided by expert knowledge, we assume the \textbf{availability of historical data} to implement automated or semi-automated methods for each of the mutually independent model components $i(\bm{x}_t)$, $p(\bm{x}_t)$, or $h(.)$. Thus, given a dataset $D \subset \mathcal{D}$, along with labels $E$ (such as the ground truth pattern type for $p$), we infer information about the state of the system at time points outside $D$. The sample should be representative of the process in both, heterogeneity and volume. The latter assumption conveys that situations, which occurred in the past, can be properly assessed, e.g. by observing consequences on the product quality. The implementation of a model component, such as the estimator for pattern type $p$, aims to generalize the knowledge obtained from $D$ to the whole data domain $\mathcal{D}$ in order to cover future outcomes. 
For this purpose, a \textbf{consistency property} ensures that knowledge from historical knowledge is correctly inferred to new data. For the pattern type classifier $p_{D,E}$, trained on information from $D$ and $E$, the property is given as
\begin{equation}
    P(\Vert p_{D,E}(\bm{x}_t) - E(\bm{x}_t) \Vert \leq \beta)\geq 1-\xi,
\end{equation}
for any $\bm{x}_t\in D$, where $\beta,\xi\geq 0$ are small constants and $\Vert .\Vert$ denotes an appropriate norm in the target space. Thus, we require that the (estimated) value of the component deviates from the ground truth labels $E$ only with small probability. For ML components, this condition is targeted by training the model on a historical dataset.
 
\subsection{Statistical properties}

% interpretation of the HF by loss functions
To understand the definition of the HFPC, we make use of statistical decision theory: the quality of a decision $d$ is modeled in the presence of an unknown state of nature $\theta$ using a loss function $L(d,\theta)\rightarrow\mathbb{R}^+$---a numerical description of the outcome that occurs if a decision $d$ is taken, while the state of the system takes a value $\theta$. In process control, the decision may represent a counter-measure to avoid production error, such as conducting additional testing or maintenance. We denote the decision to take an action at time $t$ by $d_t=1$, and $d_t=0$, otherwise. The non-observable state of nature $\bm{\theta}_t$ describes the pattern type and the intensity of the pattern at time $t$, $\bm{\theta}_t = (i(\bm{x}_t), p(\bm{x}_t), h)$. 

An intuitive expectation is that the loss in case of taking action is constant over the underlying process state since the action produces regular costs (e.g., production delays)---we summarize these costs by the threshold $\tau$. On the other hand, the loss of taking no action is modelled as proportional to both the underlying criticality of the pattern (a more critical pattern delivers a higher loss) and the intensity of the pattern (a more intense pattern delivers a higher loss). Hence, we define a loss function $L$ as follows:
\begin{equation}
    L(d_t, \bm{\theta}_t) = \left\{\begin{array}{ll}
        i(\bm{x}_t)\cdot h(p(\bm{x}_t)), & \text{if}~d_t = 0\\
        \tau, & \text{otherwise.}
    \end{array}\right.
    \label{eq:loss}
\end{equation}
It follows that the risk function $r(d_t)$, defined as the expected loss over all states of nature, is given as
\begin{equation}
    r(d_t) = \mathbb{E}_{\bm{\theta}_t}[L(d_t, \bm{\theta}_t)] =  \left\{\begin{array}{ll}
 \mathbb{E}[\text{HF}(\bm{x}_t)] & d_t=0 \\
 \tau & d_t=1
\end{array}\right.
\end{equation}
Hence, we interpret the expectation over the $\text{HF}(\bm{x}_t)$ as the risk taken, if \textbf{no action is taken}, under the loss function $L$. On the opposite, the threshold $\tau$ steers the risk of taking an action, and thus, the sensitivity of the process control system. Taking a threshold-based decision $d_t^\star$ with respect to the loss $L$ and a threshold $\tau$ is equivalent to minimizing the risk, \begin{equation}
   d_t^\star = \underset{d_t\in\{0,1\}}{\min}r(d_t).
\end{equation}

% expected failure rates
Based on this insight, we investigate the error rates associated with decisions by thresholding the HFPC: we denote the real state of the system at time $t$ by a variable $c_t\in\{0,1\}$, where $c_t=0$ denotes an uncritical state, and $c_t=1$ denotes a critical state. A pair $(d_t, c_t)$ as false positive error (FP, type I), if the system predicts an error state although no error is present, $d_t=1$ and $c_t=0$, while false negative errors (FN, type II) occur, if the system does not react to a failure, $d_t=0$ and $c_t=1$. Error rates quantify the relative frequency of occurence of the discussed error types, given as $\text{FPR}=\frac{\text{FP}}{\text{FP} + \text{TN}}$ (false positive rate) and $\text{FNR}=\frac{\text{FN}}{\text{FN} + \text{TP}}$ (false negative rate), i.e. the counts of false positives and false negatives are normalized by the counts of all (ground truth) positives or negatives, respectively, see~\cite{FAWCETT2006861}.

In the described framework, we can derive upper bounds for the error rates of $\text{HF}(\bm{x}_t)$:\footnote{for proof, see supplementary material}

\begin{theorem}[Bound for the HFPC's error rates]
    Given that intensity $i(\bm{x})$ and pattern criticality $H(\bm{x})=h(p(\bm{x}))$ are stochastically independent, the following upper bound holds for an arbitrary parameter $\tau\in[0,1]$:
    \begin{align}
        \text{FPR}_\tau &\leq (1-\tau) \cdot \text{FPR}^{(i)}_{\tau} \cdot \text{FPR}^{(H)}_{\tau}, \\
       \text{FNR}_\tau &\leq \tau,
    \end{align} 
    where $\text{FPR}^{(i)}_{\tau}$ and $\text{FPR}^{(H)}_{\tau}$ denote the false positive rates of $i(\bm{x})$ and $H(\bm{x})$, respectively.
\end{theorem}

Since the random variable $H_{\bm{x}}$ comprises two probabilistic terms, a data-driven pattern type $\bm{p}_{\bm{x}}$, and an expert judgement on the criticality $h(.)$, we aim to decompose the error rates of $H$ into its components. We define the following subsets of (ground truth) critical and uncritical pattern combinations: $\mathcal{P}_c=\{\bm{p}: h(\bm{p})>\tau\}$ and $\mathcal{P}_u=\{\bm{p}:h(\bm{p})\leq \tau\}$, leading to the following decomposition:\footnotemark[2]
\begin{theorem}[Decomposition of error rates]
    The FPR of $H$ can be decomposed as follows:
    \begin{footnotesize}
    \begin{align}
        \text{FPR}_\tau^{(H)} =  &\sum\limits_{\bm{p}\in\mathcal{P}_c} P(H(\bm{x}) > \tau | p(\bm{x}) = \bm{p}) \cdot   
        P(p(\bm{x})=\bm{p}| p(\bm{x}) \in \mathcal{P}_u) + \nonumber \\
        &\sum\limits_{\bm{p}\in\mathcal{P}_u} P(H(\bm{x}) > \tau | p(\bm{x}) = \bm{p}) \cdot P(p(\bm{x}) =\bm{p}| p(\bm{x}) \in \mathcal{P}_u),
\end{align}
\end{footnotesize}
\end{theorem}

Thus, $\text{FPR}_\tau^{(H)}$ decomposes into a sum of probabilities associated with two scenarios
\begin{enumerate}
    \item a pattern recognition error (a critical pattern is predicted erroneously),
    \item an expert assessment error (criticality of an uncritical pattern is overestimated).
\end{enumerate}
An analog decomposition holds for $\text{FNR}_\tau^{(H)}$.

We conclude that a system following (a) the definition of a process indicator HFPC, and (b) the decision process by thresholding HFPC using a threshold $\tau$, permits us to draw direct conclusions about the worst-case error rates. Thus, the user can assess the risk of making wrong decisions.

\section{Case Study from Semiconductor Industry}

In the semiconductor industry, like in many other advanced manufacturing domains, a common problem is judging the production process's stability and distinguishing between critical and uncritical types of events. While critical events can lead to violations of product specifications, uncritical events have no relevant impact on product quality. The production process consists of hundreds of synchronized process steps with limited possibilities for functional in-line measurements. Thus, the quality of the product must be judged at the end of (frontend) production, the wafer test stage, where electrical parameters are measured for each device on the wafer.
When analyzing these measurement data, characteristic spatial patterns can reveal events like process deviations.

To a large extent, state-of-the-art process monitoring in semiconductor manufacturing is based on techniques like yield loss or dynamical partial average testing (PAT)~\cite{pat}. Yield loss is a generic approach, where we take the relative number of devices per wafer, violating specification limits in the wafer test procedure as a quality indicator---the occurrence of wafers with a high number of faults triggers an alarm or counter-action in the production. Dynamical PAT is a specific version of PAT, where control limits are introduced as robust empirical quantiles of the measured electrical test values to identify and remove outliers dies on the wafer. Both methods, yield loss, and PAT cannot cover the complexity of the problem: they neither take spatial positions into account nor discriminate between critical and uncritical pattern types.

This section evaluates the Health Factor for Process Patterns (HF), introduced in our prior work~\cite{schrunner19,schrunner2019phd}, and explores the applicability of the more general concepts and insights derived in this work. In a quantitative analysis, we demonstrates that the concept is applicable and yields an improvement over the domain's state-of-the-art methods. Further, we showcase how our theoretic insights improve the understanding of the system's failure rates. \footnote{A distinct concept named \textit{Health Factor} exists for predictive maintenance~\cite{susto2015} but is unrelated to the Health Factor for Process Control discussed in this work.}

\subsection{Definition of the Health Factor}

The HF~\cite{schrunner19} for semiconductor manufacturing operates on analog wafermaps and consists of preprocessing via Markov Random Fields to remove noise and background structures, a feature extraction step, and a subsequent selection of the classification method. As image features comprising the feature space $\mathcal{D}$, a combination of Local Binary Pattern (LBP), Rotated Local Binary Pattern (RLBP) feature selectors (adapted for this purpose by Santos et al.~\cite{santos2019}) and Histogram of Oriented Gradients (HOG)~\cite{hog} is used. 
The pattern type $p(w)\in \{1,\dots,K\}$ shown on a wafermap $w\in \mathcal{D}$ is estimated via a classifier $p$. We assume that each wafermap contains only one out of $K$ pattern. The criticality $h(.)$ for each pattern type is specified via expert knowledge, such that $0$ denotes uncritical pattern types and $1$ denotes critical pattern types. The intensity $i(w)$ of wafermap $w$ ranges from weak states, where a pattern can hardly be distinguished from measurement noise to strong states, where clear gradients are present on the wafermap. While pattern types are characterized by the shape and position of the pattern on the wafermap, intensity describes the overall degree of anomalies. As suggested in ~\cite{jenul2019}, intensity is implemented via 
the Moran's I~\cite{moran1950}, denoted as $i(w)$ for wafermap $w\in \mathcal{D}$.

\subsection{Experimental setup}
Our analysis includes 3 datasets:
\begin{itemize}
\item [a)] $dataset_1$: A simulated dataset showing 5 different pattern types (1-5), comprising 400 wafermaps from each out of 5 pattern types, resulting in 2,000 wafermaps in total.\footnote{The dataset is available at \url{http://www.doi.org/10.5281/zenodo.2542504}.}
\item [b)] $dataset_2$: 4,324 wafermaps from a real-world product showing 8 different pattern types ($A$-$H$). The dataset is unbalanced due to different occurrence frequencies between types of critical events in the underlying process.
\item [c)] $dataset_3$: A dataset of a distinct semiconductor product, containing 679 wafermaps with 4 different pattern types ($\mathfrak{A}$-$\mathfrak{D}$).
\end{itemize}

In order to evaluate the Health Factor and its components, one or more critical pattern types are specified by domain experts in each of the 3 datasets. The vectors assigning criticality values to each pattern type in $dataset_1$, $dataset_2$ and $dataset_3$ are denoted by $h_1$, $h_2$ and $h_3$, respectively. For the simulated dataset $dataset_1$, the criticality is set to $h_1(1) = 1, h_1(2) = h_1(4) = 0.5, h_1(3) = h_1(5) = 0$. For $dataset_2$ the critical pattern type is $H$, while for $dataset_3$, pattern type $\mathfrak{A}$ is critical, such that $h_2(H) = 1$ and $h_3(\mathfrak{A})=1$. All other pattern types get assigned criticality $0$.

The evaluation procedure is conducted in 3 different setups:
a) evaluation of the pattern intensity concept, compared with state-of-the-art concepts yield loss and dynamical PAT (control charts), b) evaluation of the pattern type classification concept, c) evaluation of the Health Factor thresholds.
Each setup is repeated 100 times on each dataset.
In each run, balanced train and test subsets are sampled: the train-test-splits of $dataset_1$ consist of 100 wafermaps (per pattern type) in the training and 100 wafermaps (per pattern type) in the test set (100/100). In $dataset_2$, a train-test-split of 100/147 and in $dataset_3$, a train-test-split of 40/45 is chosen.

\begin{figure}
    \centering
    \begin{subfigure}{0.45\textwidth}
        \centering
		\includegraphics[width = 0.25\textwidth]{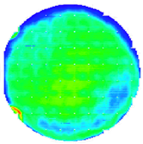}
		\includegraphics[width = 0.25\textwidth]{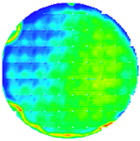}
		\includegraphics[width = 0.25\textwidth]{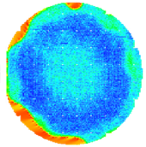}
		\caption{Wafermaps showing weak (left), mediocre (middle) and strong (right) intensities of a single pattern $H$.} 
		\label{fig:results_experiment2a}
	\end{subfigure}
	\begin{subfigure}{0.45\textwidth}
	    \centering
		\includegraphics[width = 1.0\textwidth]{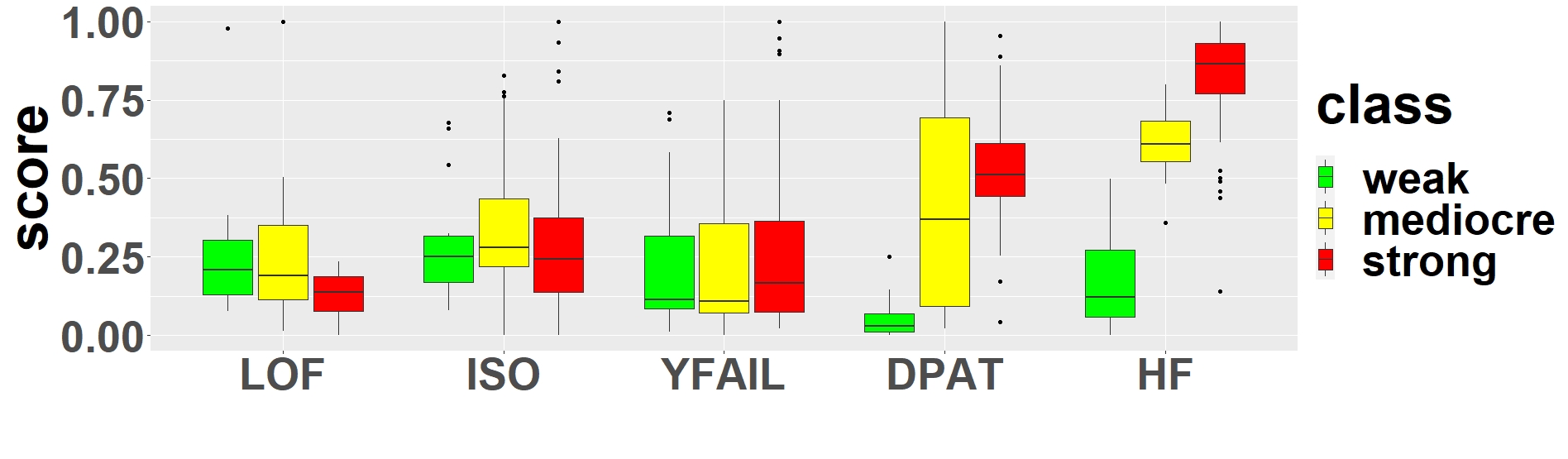}
		\caption{Boxplots of intensity values from different indicators.}
		\label{fig:results_experiment2b}
	\end{subfigure}
    \caption{Intensity quantification results comparing Health Factor (HF), Local Outlier Factor (LOF), Isolation Forest (ISO), Yield Loss (YFAIL), and dynamical PAT (DPAT). Coloring of the boxplots represents expert judgment, with HF separating all three of intensity classes.}
    \label{fig:results_experiment2}
\end{figure}

\subsection{Results}

In our initial evaluation, we focus on intensity variation within a (known) critical pattern class of $dataset_2$ --- this aspect is a direct extension to concepts like yield loss and PAT. Therefore, we leave the classification component out and restrict it to the subset of wafermaps from $dataset_2$, which are assigned to pattern type $H$ (a total number of $170$ wafermaps). These wafermaps are manually classified into three categories representing the degree of development of the pattern, shown in Fig. \ref{fig:results_experiment2a}: 
\begin{itemize}
    \item weak (73 wafermaps): slight pattern contours,
    \item mediocre (20 wafermaps): contours visible with low contrast,
    \item strong (77 wafermaps): clear pattern contours.
\end{itemize}
We analyze the Health Factor (HF), dynamical PAT (DPAT), and yield loss (YFAIL) values delivered for each wafermap concerning these ground truth categories. Additionally, we compare our results to two commonly used anomaly detectors, namely the Local Outlier Factor (LOF) and Isolation Forests (ISO). The final anomaly scores for LOF and ISO are the highest (=worst) anomaly scores overall investigated electrical tests performed on each wafer to keep the comparison as fair as possible. Results are presented as boxplots in Fig. \ref{fig:results_experiment2b}: HF discriminates well between all three categories with few outliers, while DPAT values merely differ between wafermaps with patterns of weak and mediocre/strong intensities, and all other methods cannot cover distinct intensity levels accurately.

\begin{figure}
\begin{subfigure}{0.55\textwidth}
    \centering
    \includegraphics[width=\textwidth]{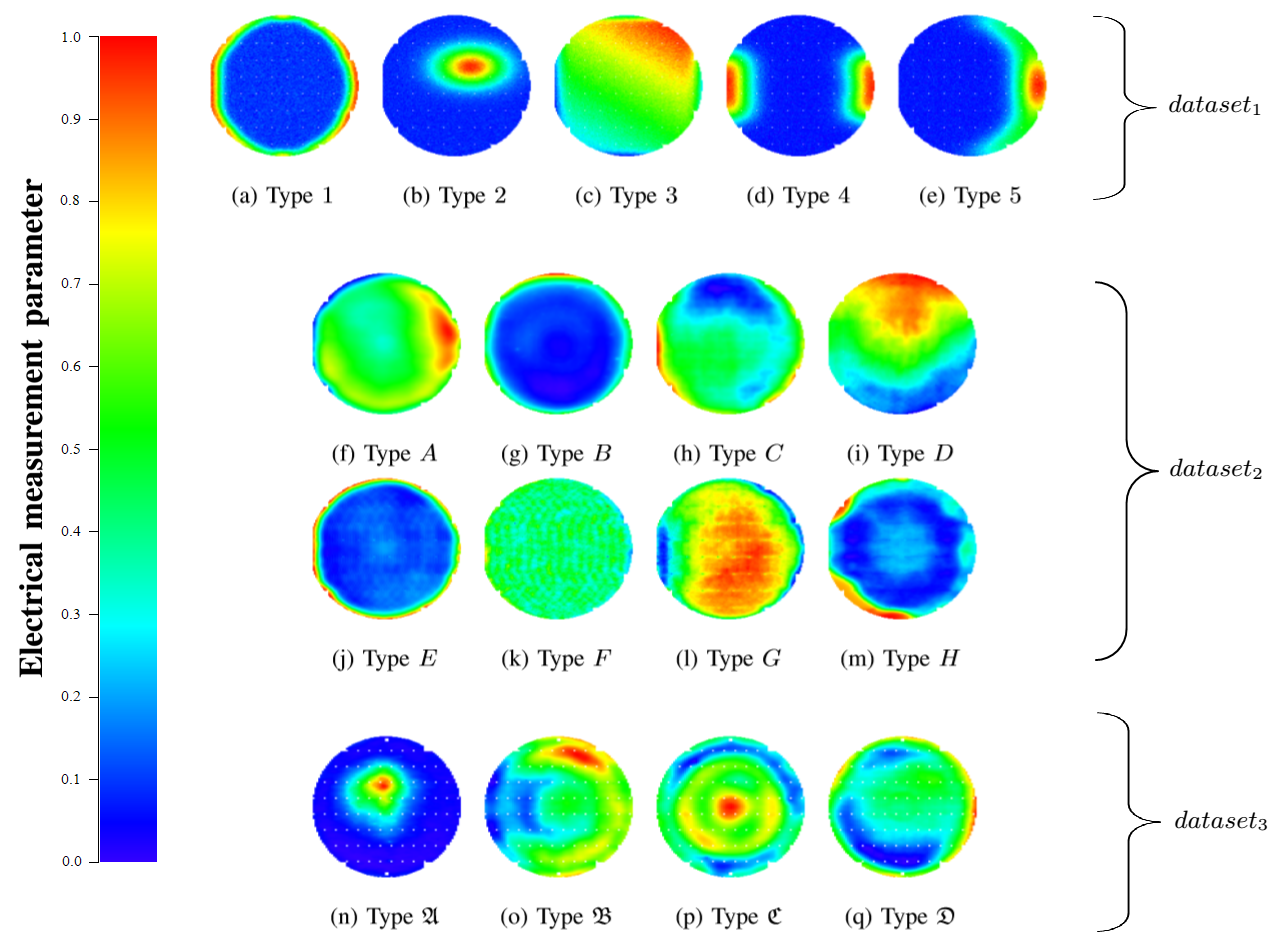}
    \caption{Example wafermaps of pattern types: wafermaps (a)-(e) correspond to $dataset_1$, wafermaps (f)-(m) to $dataset_2$ and wafermaps (n)-(q) to $dataset_3$. The values range from low (blue) to high (red).}
    \label{fig:wafermaps}
\end{subfigure}\hfill
\begin{subfigure}{0.4\textwidth}
    \resizebox{0.95\textwidth}{!}{
    \begin{tabular}{|c|c|r|l|l|l|l|l|}
    \cline{4-8}
    \multicolumn{3}{c|}{} & \multicolumn{5}{|c|}{F1 score} \\ \cline{2-8}
    \multicolumn{1}{c|}{}& pattern & \multirow{2}{*}{total} & \multirow{2}{*}{$NB$} & \multirow{2}{*}{$RF$} & \multirow{2}{*}{$SVM_l$} & \multirow{2}{*}{$SVM_p$} & \multirow{2}{*}{$LR$} \\ 
    \multicolumn{1}{c|}{}& type & & & & & & \\\cline{1-8}
    \multirow{6}{*}{\rotatebox[origin=c]{90}{$dataset_1$}} {} & 1 & 400 & \textbf{1.000} & \textbf{1.000} & \textbf{1.000} & 0.867 & \textbf{1.000} \\ \cline{2-8}
{} & 2 & 400 & 0.994 & \textbf{1.000} & 0.964 & 0.775 & 0.999 \\ \cline{2-8}
{} & 3 & 400 & \textbf{1.000} & \textbf{1.000} & \textbf{1.000} & 0.978 & \textbf{1.000}  \\ \cline{2-8}
{} & 4 & 400 & \textbf{1.000} & \textbf{1.000} & 0.981 & 0.794 & 0.999 \\ \cline{2-8}
{} & 5 & 400 & 0.994 & \textbf{1.000} & 0.984 & 0.959 & \textbf{1.000} \\ \cline{2-8}
{} &\multicolumn{2}{|c|}{average} & 0.997 & \textbf{1.000} & 0.986 & 0.874 & \textbf{1.000} \\ \hline     \hline
    \multirow{9}{*}{\rotatebox[origin=c]{90}{$dataset_2$}} {} &  $A$ & 568 & 0.998 & \textbf{1.000} & \textbf{1.000} & 0.933 & 0.999 \\ \cline{2-8}
{} & $B$ & 566 & 0.994 & 0.994 & 0.965 & 0.728 & \textbf{0.997} \\ \cline{2-8}
{} & $C$ & 918 & 0.868 & 0.997 & 0.943 & 0.403 & \textbf{0.998} \\ \cline{2-8}
{} & $D$ & 568 & 0.981 & 0.990 & 0.969 & 0.788 & \textbf{0.994} \\ \cline{2-8}
{} & $E$ & 284 & \textbf{1.000} & \textbf{1.000} & 0.997 & 0.990 & \textbf{1.000} \\ \cline{2-8}
{} & $F$ & 852 & 0.975 & 0.995 & 0.979 & 0.956 & \textbf{0.996} \\ \cline{2-8}
{} & $G$ & 284 & 0.979 & \textbf{0.993} & 0.959 & 0.807 & \textbf{0.993} \\ \cline{2-8}
{} & $H$ & 284 & 0.807 & 0.990 & 0.910 & 0.675 & \textbf{0.997} \\ \cline{2-8}
{} &\multicolumn{2}{|c|}{average}  & 0.950 & 0.995 & 0.965 & 0.785 & \textbf{0.997} \\ \hline \hline
\multirow{5}{*}{\rotatebox[origin=c]{90}{$dataset_3$}} {} &  $\mathfrak{A}$ & 194 & 0.998 & \textbf{1.000} & \textbf{1.000} & 0.991 & 0.932 \\ \cline{2-8}
{} & $\mathfrak{B}$ & 194 & 0.935 & \textbf{0.970} & 0.870 & 0.709 & 0.879 \\ \cline{2-8}
{} & $\mathfrak{C}$ & 194 & 0.922 & \textbf{0.965} & 0.873 & 0.783 & 0.864 \\ \cline{2-8}
{} & $\mathfrak{D}$ & 97 & 0.976 & \textbf{0.986} & 0.954 & 0.921 & 0.900 \\ \cline{2-8}
{} &\multicolumn{2}{|c|}{average} & 0.958 & \textbf{0.980} & 0.924 & 0.851 & 0.894 \\ \hline
    \end{tabular}
}
    \caption{Macro-averaged F1 scores for each dataset, pattern type and classifier.}
    \label{fig:simResults}
\end{subfigure}
\caption{Experimental data and pattern type classification results on 3 datasets.}
\label{fig:classification}
\end{figure}

A one-way analysis of variance (ANOVA) is performed to investigate whether any of the five indicators HF, DPAT, YFAIL, ISO, and LOF yields significant group differences with respect to groups weak, mediocre, and strong. 
ISO and YFAIL do not indicate the existence of substantial group differences (p-values $>10^{-1}$), whereas low p-values ($<10^{-5}$) for LOF, HF, and DPAT indicate group differences. In contrast to DPAT and LOF, the pairwise $t$-test with Tukey correction for simultaneous testing confirms that HF values significantly differ between all pairs of groups (p-values $<10^{-5}$). 
Hence, ANOVA supports the observation that HF is better suited to discriminate between weak, mediocre, and strong patterns on wafermaps. At the same time, DPAT can detect weak patterns and strong patterns can be detected by LOF, with similar performance. ISO and YFAIL fail to indicate class differences accurately.

In order to exploit the full power of ML concepts, we focus now on the distinction between pattern types: we evaluate five distinct classification methods to predict the pattern type based on the wafermap features: a) Naive Bayes classifier ($NB$), b) Random Forest ($RF$), c) Support Vector Machine (SVM) with linear kernel ($SVM_l$), d) SVM with polynomial kernel ($SVM_p$), e) One-vs-one Logistic Regression ($LR$).
Classification quality is evaluated via F1 scores, representing the harmonic mean of precision and recall for each pattern type. To address the multi-class problem, the F1 score is averaged over all classes (macro-averaging). Fig. \ref{fig:classification} presents sample wafermaps from all classes, along with calculated F1 scores for each dataset, pattern type and classification method. All classifiers yield highly accurate results, underlining that the set of features covers all necessary information to model the pattern type. Except for $SVM_p$, all classification methods achieve an F1 score of at least $0.90$ for all three datasets. $RF$ classification yields the best results with an average F1 score of at least $0.98$. On the other hand, $LR$ seems to outperform the $RF$ for larger datasets.

\begin{figure*}
	\centering
	\begin{subfigure}{0.23\textwidth}
		\includegraphics[width=\textwidth]{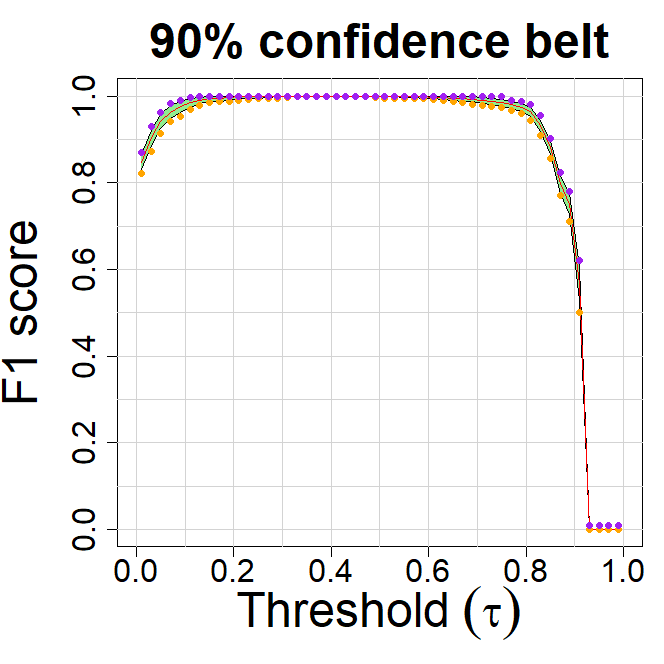}
		\caption{$dataset_1$}
		\label{fig:results_experiment3a}
    \end{subfigure}
	\begin{subfigure}{0.23\columnwidth}
		\includegraphics[width=\textwidth]{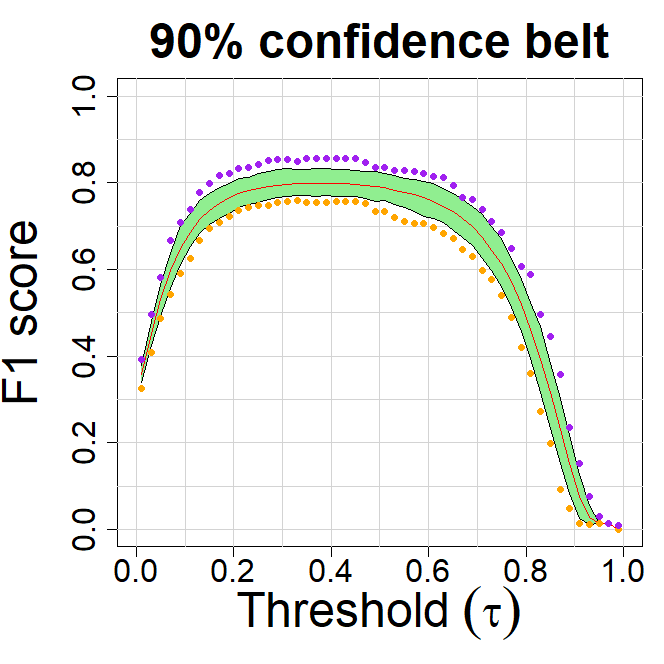}
		\caption{$dataset_2$}
		\label{fig:results_experiment3b}
	\end{subfigure}
	\begin{subfigure}{0.23\columnwidth}
		\includegraphics[width=\textwidth]{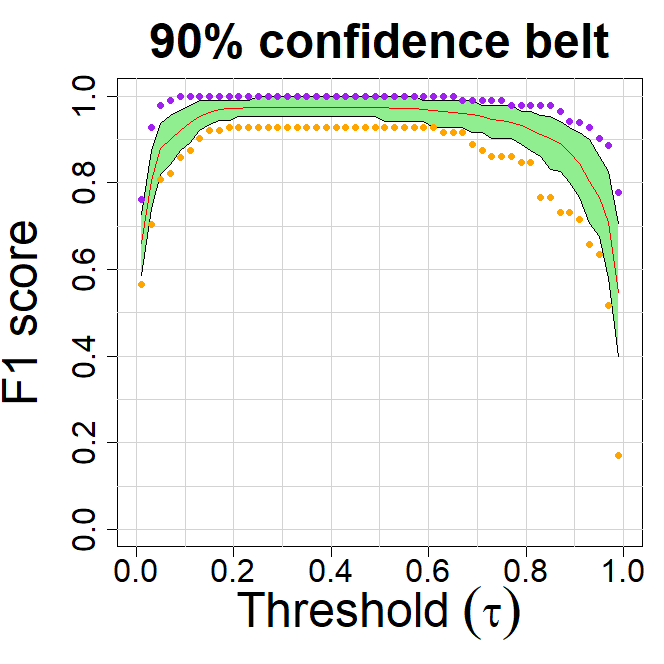}
		\caption{$dataset_3$}
		\label{fig:results_experiment3c}
	\end{subfigure}
	\begin{subfigure}{0.23\columnwidth}
		\resizebox{\textwidth}{!}{
		\begin{tabular}{c|c|c|c}
			Fig. & $\tau$ & mean & sd \\ \hline
			\ref{fig:results_experiment3a} & 0.15 & 0.993 & 0.004 \\
			\ref{fig:results_experiment3b} & 0.35 & 0.798 & 0.018\\
			\ref{fig:results_experiment3c} & 0.09 & 0.921 & 0.032 \\
		\end{tabular}	
		}
		\caption{Optimal threshold $t$ (parameter, where the highest mean value was obtained), as well as mean and standard deviation over 100 runs.} 
	\end{subfigure}
	\caption{$90\%$ confidence belts of averaged F1 scores for each dataset of the F1 score depending on the chosen threshold. Minimum and maximum values per threshold are indicated as purple and orange dots, respectively.}
	\label{fig:results_experiment3}
\end{figure*}

As a final experiment, we evaluate the entire concept of the HF at once. As a ground truth for the predicted HF values, the expert assigns a 0-1 coding to each wafermap, quantifying whether the wafermap depicts a critical pattern, which should trigger an alarm. 
F1 scores are calculated for each dataset when comparing the ground truth 0-1-coding to the thresholded HF value with threshold $\tau$. Pattern type classification is based on $RF$ due to its superior overall performance in the previous experiment. Fig. \ref{fig:results_experiment3} shows the resulting F1 scores averaged over $100$ runs, along with $90\%$ confidence belts, as function of $\tau$. The results indicate that the F1 scores are robust concerning selecting the threshold $\tau$, which is relevant for practical applications. The maximum F1 score for $dataset_2$ of approx. $0.8$ is lower than for $dataset_1$ and $dataset_3$, which suggests that the HF (but probably also the definition of accurate ground truth values) suffers from a growing number of pattern types.

\begin{figure*}
    \centering
    \begin{subfigure}{0.45\textwidth}
        \centering
        \includegraphics[width = \textwidth]{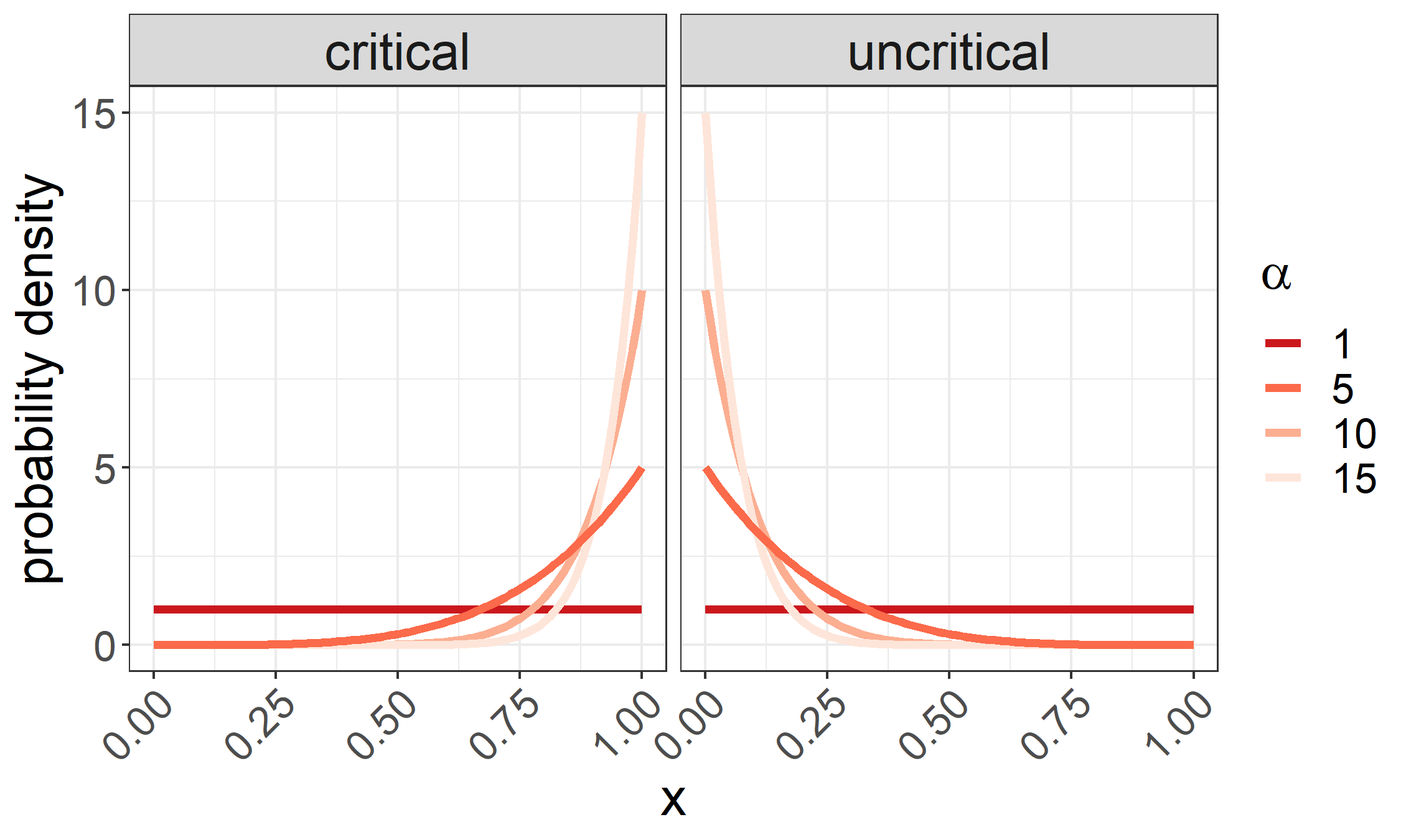}
        \caption{Beta distributions representing non-deterministic criticality.}
        \label{fig:results_experiment_4a}
    \end{subfigure}
    \begin{subfigure}{0.45\textwidth}
        \centering
        \includegraphics[width = \textwidth]{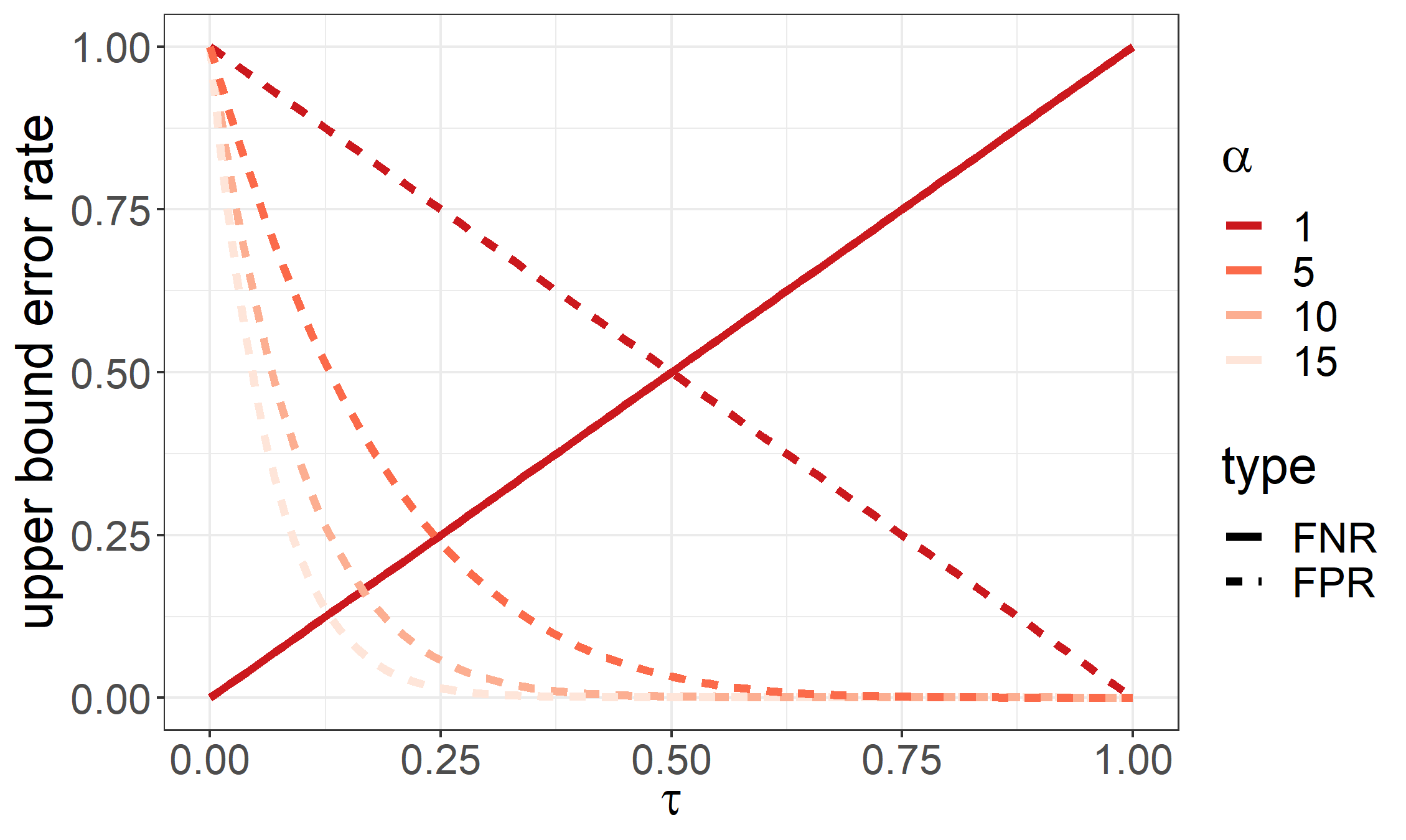}
        \caption{Upper bounds for FPR and FNR.}
        \label{fig:results_experiment_4b}
    \end{subfigure}
    \caption{Upper bounds for HFPC error rates at different levels of uncertainty associated with criticality, modeled via Beta distributions with different shape parameters $\alpha$. The strong impact of $\alpha$ illustrates the importance of expert knowledge: if experts can precisely assess the criticality, $\tau$ is more robust and HFPC errors are reduced.
    }
\end{figure*}

In the given situation, our theoretical results allow us to quantify the upper bounds for failure rates, FPR, and FNR. The confusion matrix of the SVM classifier evaluated on the test data of $dataset_2$ delivers error probabilities of the pattern recognition component $p$. Expert knowledge on the pattern criticality $h_2$ is modelled by a Beta distribution $h_2(\text{H})=h_\text{crit}\sim Be(\alpha, 1)$ for critical class $H$, and $h_2(i) = h_\text{uncrit} \sim \text{Be}(1,\alpha)$ for all uncritical classes $i\neq \text{H}$ with uncertainty represented by shape parameter $\alpha>1$, as shown in Fig. \ref{fig:results_experiment_4a}. For simplicity, intensity is assumed to be deterministic. Fig. \ref{fig:results_experiment_4b} shows the estimated upper bounds for FPR and FNR, see Section \ref{sec:theory}. Since intensities are assumed to be deterministic, and pattern types are distinguished with high accuracy, error estimates depend mainly on the uncertainty associated with the criticality component. When increasing $\alpha$, indicating that the expert's assessment of criticality levels becomes more reliable, a decrease in the upper bound of FPR is observed for small threshold parameters $\tau$---thus, the system is less sensitive to the parameter selection for $\tau$. On the contrary, if $\alpha$ is high, larger error rates are expected, and thus, an appropriate selection of $\tau$ becomes essential.

\section{Discussion and Conclusion}
We presented Health Factor for Process Control (HFPC), a framework to design indicators for process monitoring in automated production systems, which allows integration of pattern recognition and ML methods based on clear assumptions. We analyzed the statistical properties of the proposed Health Factor, demonstrating how errors in the framework can be traced back to the errors of its single components. For ML-based components, the respective errors can be estimated via sample test data. 
We analyzed one application of the proposed setup in a case study. We evaluated the performance of the Health Factor on three datasets from the semiconductor industry and verified that the concept has theoretical and practical value.
We found the reliability of expert knowledge to be critical for the influence of the error rates of the ML method on the overall risk. 

According to its mathematical properties, the HFPC is guaranteed to be reasonable from the viewpoint of human understanding, enabling automated or semi-automated actions to be taken. The proposed framework can simplify the transition from purely statistical key numbers to ML-based approaches in practical applications beyond the application in the semiconductor industry. Adapting the concept to other industries is feasible since the concept is exclusively dependent on the input data structure and expert knowledge (and not the source or format of the data). For instance, the framework can also be applied to time-series data if the feature extraction step is tailored. Further, the flexibility to integrate challenging ML scenarios in the process monitoring framework, such as incremental learning~\cite{kong20} or the presence of unknown classes \cite{schrunner20}, is a substantial benefit.

\section*{Acknowledgment}
A part of the work has been performed in the project Power Semiconductor and Electronics Manufacturing 4.0 (SemI40), under grant agreement No 692466. The project is co-funded by grants from Austria (BMVIT-IKT der Zukunft, FFG project no. 853338 and 6053321), Germany, Italy, France, Portugal and - Electronic Component Systems for European Leadership Joint Undertaking (ECSEL JU).

Further, a part of the work has been performed in the project Arrowhead Tools for Engineering of Digitalisation Solutions (Arrowhead Tools). This project has been funded by the European Commission, through the European H2020 research and innovation programme, ECSEL Joint Undertaking, and National Funding Authorities from 18 involved countries under the research project Arrowhead Tools with Grant Agreement no. 826452.

\newpage
\section*{Proofs}
\paragraph{Notations}
In the following, we denote probability densities of continuous random variables by $f_X(x)$, where $X$ denotes the random variable and $x$ denotes the realization (analogous to $P(X=x)$ in a discrete setting). Conditional densities are denoted by $f_{X|Y=y}(x)$. 

\begin{proof}[Proof of Theorem 1]
    The claim is directly derived as follows:
    \begin{align}
        \text{FPR}_{\tau}  &= P(i(\bm{x})\cdot H(\bm{x}) > \tau | c_t = 0) \\
        &= \int\limits_{\tau}^{1}f_{i(\bm{x})\cdot H(\bm{x})| c_t = 0}(\rho) d\rho \\
        &\underset{(a)}{=} \int\limits_{\tau}^{1}\int\limits_{0}^{1}\int\limits_{0}^{1} \overbrace{f_{i(\bm{x})\cdot H(\bm{x})|i(\bm{x})=i,H(\bm{x})=H,i(\bm{x})\cdot H(\bm{x})\leq \tau}(\rho)}^{=\left\{
        \begin{array}{ll}
            1 & \text{if}~i\cdot H > \rho \\
            0 & \text{otherwise}\\
        \end{array}\right.
        } \cdot \nonumber \\
        & ~~~~~~~~~~~~~~
        f_{i(\bm{x}),H(\bm{x})|c_t = 0}(i,H) ~dH~di~d\rho \\
        &\underset{(b)}{=} \int\limits_{\tau}^{1}\int\limits_{\rho}^{1}\int\limits_{\frac{\rho}{i}}^{1} \underbrace{f_{i(\bm{x}), H(\bm{x})c_t = 0}(i,H)}_{\geq 0} ~dH~di~d\rho \\
        &\underset{\frac{\rho}{i}\geq \rho}{\leq}                         \int\limits_{\tau}^{1}\int\limits_{\rho}^{1}\int\limits_{\rho}^{1} f_{i(\bm{x}), H(\bm{x})|c_t = 0}(i,H) ~dH~di~d\rho \\
        &\underset{(c)}{=} \int\limits_{\tau}^{1}\int\limits_{\rho}^{1}\int\limits_{\rho}^{1} f_{i(\bm{x})|c_t = 0}(i) \cdot f_{H(\bm{x})|c_t = 0}(H) ~dH~di~d\rho \\
        &= \int\limits_{\tau}^{1}\int\limits_{\rho}^{1}f_{i(\bm{x})|c_t = 0}(i) ~di \cdot \int\limits_{\rho}^{1} f_{H(\bm{x})|c_t = 0}(H) ~dH~d\rho \\
        &= \int\limits_{\tau}^{1}\text{FPR}_\rho^{(i)} \cdot \text{FPR}_\rho^{(H)}~d\rho \\
        &\leq (1-\tau) \cdot \underset{\rho\in[\tau,1]}{\max} \left(\text{FPR}_{\rho}^{(i)} \cdot \text{FPR}_{\rho}^{(H)}\right) \\
        &\underset{(d)}{=} (1-\tau) \cdot \text{FPR}_{\tau}^{(i)} \cdot \text{FPR}_{\tau}^{(H)}.
    \end{align}
    \begin{itemize}
        \item[(a)] we use the law of total probability in the scheme $P(A|B)=\int P(A|B,C)\cdot P(C|B)dC$
        \item[(b)] the condition $i\cdot H>\rho$ with $i,H,\rho\in [0,1]$ is equivalent to $i\in(\rho,1]$ and $H\in(\frac{\rho}{i},1]$
        \item[(c)] independence of $i(\bm{x})$ and $H(\bm{x})$
        \item[(d)] both, $\text{FPR}_{\rho}^{(i)}$ and $\text{FPR}_{\rho}^{(H)}$ are monotonically decreasing in $\rho$
    \end{itemize}
    
    Similar to the false positive case, we see that
    \begin{align}
        \text{FNR}_{\tau} &= P(i(\bm{x})\cdot H(\bm{x}) \leq \tau | c_t = 1) \\
        &= \int\limits_{0}^{\tau}f_{i(\bm{x})\cdot H(\bm{x})|c_t = 1}(\rho) d\rho \\
        &= \int\limits_{0}^{\tau}\int\limits_{0}^{1}\int\limits_{0}^{1} \overbrace{f_{i(\bm{x})\cdot H(\bm{x})|i(\bm{x})=i,H(\bm{x})=H,i(\bm{x})\cdot H(\bm{x}) > \tau}(\rho)}^{=\left\{
        \begin{array}{ll}
            1 & \text{if}~i\cdot H \leq \rho \\
            0 & \text{otherwise}\\
        \end{array}\right.
        } \cdot \\
        & ~~~~~~~~~~~~~~ f_{i(\bm{x}),H(\bm{x})|c_t = 1}(i,H) ~dH~di~d\rho \\
        &\underset{(a)}{=} \int\limits_{0}^{\tau}\int\limits_{0}^{1}\int\limits_{0}^{\min\{\frac{\rho}{i},1\}} f_{i(\bm{x}),H(\bm{x})|c_t = 1}(i,H) ~dH~di~d\rho \\
        &\underset{(b)}\leq \int\limits_{0}^{\tau}\underbrace{\int\limits_{0}^{1}\int\limits_{0}^{1} f_{i(\bm{x}),H(\bm{x})|c_t = 1}(i,H) ~dH~di}_{=1}~d\rho \\
        &= \int\limits_{0}^{\tau} 1 d\rho = \tau
    \end{align}
    \begin{itemize}
        \item[(a)] the condition $i\cdot H\leq\rho$ with $i,H,\rho\in [0,1]$ is equivalent to $i\in[0,1]$ and $H\in[0,\frac{\rho}{i}] \cap [0,1] = [0,\min\{\frac{\rho}{i},1\}]$
        \item[(b)] $\min\{\frac{\rho}{i},1\}\leq 1$
    \end{itemize}
\end{proof}

\begin{proof}[Proof of Theorem 2]
\begin{align}
       \text{FPR}_\tau^{(H)} &= P(H(\bm{x}) > \tau |  c_t = 0) \\
        &= P(H(\bm{x}) > \tau | c_t = 0) \\
        &= \sum\limits_{\bm{p}\in\mathcal{P}} P(H(\bm{x}) > \tau | p(\bm{x}) = \bm{p}, c_t = 0) \cdot P(p(\bm{x})=\bm{p} | c_t = 0) \\
        &= \sum\limits_{\bm{p}\in\mathcal{P}_c} P(H(\bm{x}) > \tau | p(\bm{x}) = \bm{p}) \cdot P(p(\bm{x})= \bm{p} | c_t = 0)  + \nonumber \\
        & ~~~ \sum\limits_{\bm{p}\in\mathcal{P}_u} P(H(\bm{x}) > \tau | p(\bm{x}) = \bm{p}) \cdot P(p(\bm{x})= \bm{p} | c_t = 0) .
\end{align}
\end{proof}

% references section
\bibliographystyle{unsrt} 
\bibliography{literature}

\begin{thebibliography}{10}

\bibitem{shewhart1931}
W.~A. Shewhart.
\newblock {\em Economic Control of Quality of Manufactured Product}.
\newblock D. Van Nostrand Company, Inc., New York, 1931.

\bibitem{yu2012}
J.~{Yu}.
\newblock Semiconductor manufacturing process monitoring using gaussian mixture
  model and bayesian method with local and nonlocal information.
\newblock {\em IEEE Transactions on Semiconductor Manufacturing},
  25(3):480--493, Aug 2012.

\bibitem{liu2020}
Y.~Liu, J.~Zeng, J.~Bao, and L.~Xie.
\newblock A unified probabilistic monitoring framework for multimode processes
  based on probabilistic linear discriminant analysis.
\newblock {\em IEEE Transactions on Industrial Informatics}, 16(10):6291--6300,
  2020.

\bibitem{GAN202283}
Chao Gan, Wei-Hua Cao, Kang-Zhi Liu, and Min Wu.
\newblock A novel dynamic model for the online prediction of rate of
  penetration and its industrial application to a drilling process.
\newblock {\em Journal of Process Control}, 109:83--92, 2022.

\bibitem{7560618}
Jia~Peter Liu, Omer~F. Beyca, Prahalad~K. Rao, Zhenyu~James Kong, and Satish
  T.~S. Bukkapatnam.
\newblock Dirichlet process gaussian mixture models for real-time monitoring
  and their application to chemical mechanical planarization.
\newblock {\em IEEE Transactions on Automation Science and Engineering},
  14(1):208--221, 2017.

\bibitem{7556960}
Mohammad~Samie Tootooni, Prahalad~K. Rao, Chun-An Chou, and Zhenyu~James Kong.
\newblock A spectral graph theoretic approach for monitoring multivariate time
  series data from complex dynamical processes.
\newblock {\em IEEE Transactions on Automation Science and Engineering},
  15(1):127--144, 2018.

\bibitem{ZHANG2020173}
Chuanfang Zhang, Kaixiang Peng, and Jie Dong.
\newblock An incipient fault detection and self-learning identification method
  based on robust svdd and rbm-pnn.
\newblock {\em Journal of Process Control}, 85:173--183, 2020.

\bibitem{SUN202030}
Bei Sun, Chunhua Yang, Yalin Wang, Weihua Gui, Ian Craig, and Laurentz Olivier.
\newblock A comprehensive hybrid first principles/machine learning modeling
  framework for complex industrial processes.
\newblock {\em Journal of Process Control}, 86:30--43, 2020.

\bibitem{DOMAHIDI2014763}
Alexander Domahidi, Fabian Ullmann, Manfred Morari, and Colin~N. Jones.
\newblock Learning decision rules for energy efficient building control.
\newblock {\em Journal of Process Control}, 24(6):763--772, 2014.
\newblock Energy Efficient Buildings Special Issue.

\bibitem{schrunner19}
Stefan Schrunner, Anna Jenul, Michael Scheiber, Anja Zernig, Andre Kaestner,
  and Roman Kern.
\newblock A health factor for process patterns enhancing semiconductor
  manufacturing by pattern recognition in analog wafermaps.
\newblock In {\em 2019 IEEE International Conference on Systems, Man and
  Cybernetics (SMC)}, pages 3555--3560, 2019.

\bibitem{FAWCETT2006861}
Tom Fawcett.
\newblock An introduction to roc analysis.
\newblock {\em Pattern Recognition Letters}, 27(8):861--874, 2006.
\newblock ROC Analysis in Pattern Recognition.

\bibitem{pat}
{Automotive Electronics Council}.
\newblock Guidelines for part average testing, Dec 2011.
\newblock accessed March 11, 2022.

\bibitem{schrunner2019phd}
Stefan Schrunner.
\newblock {\em Pattern Recognition in Analog Wafer Test Data - A Health Factor
  for Process Patterns}.
\newblock PhD thesis, Graz University of Technology, Graz, Austria, 08 2019.

\bibitem{susto2015}
G.~A. {Susto}, A.~{Schirru}, S.~{Pampuri}, S.~{McLoone}, and A.~{Beghi}.
\newblock Machine learning for predictive maintenance: A multiple classifier
  approach.
\newblock {\em IEEE Transactions on Industrial Informatics}, 11(3):812--820,
  Jun 2015.

\bibitem{santos2019}
T.~Santos, S.~Schrunner, B.~C. Geiger, O.~Pfeiler, A.~Zernig, A.~Kaestner, and
  R.~Kern.
\newblock Feature extraction from analog wafermaps: A comparison of classical
  image processing and a deep generative model.
\newblock {\em IEEE Transactions on Semiconductor Manufacturing},
  32(2):190--198, May 2019.

\bibitem{hog}
N.~Dalal and B.~Triggs.
\newblock Histograms of oriented gradients for human detection.
\newblock In {\em IEEE Computer Society Conference on Computer Vision and
  Pattern Recognition (CVPR'05)}, pages 886--893, June 2005.

\bibitem{jenul2019}
Anna Jenul.
\newblock Intensity quantification of process patterns in wafer test data.
\newblock Master's thesis, {Alpen-Adria-Universität Klagenfurt}, Apr 2019.

\bibitem{moran1950}
P.~A.~P. Moran.
\newblock Notes on continuous stochastic phenomena.
\newblock {\em Biometrika}, 37(1):17--23, Jun 1950.

\bibitem{kong20}
Y.~Kong and D.~Ni.
\newblock A semi-supervised and incremental modeling framework for wafer map
  classification.
\newblock {\em IEEE Transactions on Semiconductor Manufacturing}, 33:62--71,
  2020.

\bibitem{schrunner20}
Stefan Schrunner, Bernhard~C. Geiger, Anja Zernig, and Roman Kern.
\newblock A generative semi-supervised classifier for datasets with unknown
  classes.
\newblock In {\em Proceedings of the 35th Annual ACM Symposium on Applied
  Computing}, SAC '20, page 1066–1074, New York, NY, USA, 2020. Association
  for Computing Machinery.

\end{thebibliography}

\end{document}